\documentclass[conference]{IEEEtran}
\IEEEoverridecommandlockouts
\usepackage{cite}
\usepackage{amsmath,amssymb,amsfonts}
\usepackage{algorithmic}
\usepackage{graphicx}
\usepackage{textcomp}
\usepackage{xcolor}
\usepackage[colorlinks,
            linkcolor=blue,
            anchorcolor=blue,
            citecolor=blue]{hyperref}
\usepackage{stfloats}
\usepackage{booktabs}
\def\BibTeX{{\rm B\kern-.05em{\sc i\kern-.025em b}\kern-.08em
    T\kern-.1667em\lower.7ex\hbox{E}\kern-.125emX}}
\begin{document}

\title{SAM4Dcap: Training-free Biomechanical Twin System from Monocular Video\\
\thanks{This study was supported by Key Research Project of Science \& Technology Department of Sichuan Province, China (2024YFFK0041) and YunDan Health Mangement Project (3122330792)}
}

% \author{\IEEEauthorblockN{1\textsuperscript{st} Dongjie Cheng}
% \IEEEauthorblockA{\textit{West China Biomedical Big Data Center,}
% \\ \textit{West China Hospital, Sichuan University} \\
% Chengdu, China \\
% chengdongjie@stu.scu.edu.cn}
% \and
% \IEEEauthorblockN{2\textsuperscript{nd} Ziyuan Qin}
% \IEEEauthorblockA{\textit{West China Biomedical Big Data Center,}
% \\ \textit{West China Hospital, Sichuan University} \\
% Chengdu, China}
% \and
% \IEEEauthorblockN{2\textsuperscript{nd} Zekun Jiang}
% \IEEEauthorblockA{\textit{West China Biomedical Big Data Center,}
% \\ \textit{West China Hospital, Sichuan University} \\
% Chengdu, China}
% \and
% \IEEEauthorblockN{3\textsuperscript{rd} Shaoting Zhang}
% \IEEEauthorblockA{\textit{Shanghai Artificial Intelligence Laboratory}
%  \\
% Shanghai, China}
% \and
% \IEEEauthorblockN{\IEEEauthorrefmark{2} Qicheng Lao \thanks{* Qicheng Lao is corresponding author}}
% \IEEEauthorblockA{\textit{School of Artificial Intelligence, BUPT}
%  \\
% Beijing, China\\
% qicheng.lao@bupt.edu.cn}
% \and
% \IEEEauthorblockN{\IEEEauthorrefmark{2} Kang Li \thanks{* Kang Li is corresponding author}}
% \IEEEauthorblockA{\textit{West China Biomedical Big Data Center,}
% \\ \textit{West China Hospital, Sichuan University} \\
% Chengdu, China \\
% likang@wchscu.cn}

\author{
    \IEEEauthorblockN{
        Li Wang\IEEEauthorrefmark{2}\IEEEauthorrefmark{1},
        HaoYu Wang\IEEEauthorrefmark{3}\IEEEauthorrefmark{1},
        Xi Chen\IEEEauthorrefmark{2},
        ZeKun Jiang\IEEEauthorrefmark{3}\IEEEauthorrefmark{7},
        Kang Li\IEEEauthorrefmark{3}\IEEEauthorrefmark{7},
        Jian Li\IEEEauthorrefmark{2}\IEEEauthorrefmark{7}
    }
    
    \IEEEauthorblockA{\IEEEauthorrefmark{2}Sports Medicine Center, West China Hospital, West China School of Medicine, Sichuan University, Chengdu, Sichuan, China}
    
    \IEEEauthorblockA{\IEEEauthorrefmark{3}West China Biomedical Big Data Center, West China Hospital, Sichuan University, Chengdu, China}
    
    \IEEEauthorblockA{
        \IEEEauthorrefmark{1}Equal contribution: Li Wang, HaoYu Wang, 
        \IEEEauthorrefmark{7}Corresponding Author: ZeKun Jiang, Kang Li, Jian Li
    }
    
    \IEEEauthorblockA{
        Email: wangli1@stu.scu.edu.cn, why\_box@163.com, geteff@wchscu.edu.cn, \\
        zekun\_jiang@163.com, likang@wchscu.cn, lijian\_sportsmed@163.com
    }
}

% \and
% \and
% \and

\maketitle

\begin{abstract}
Quantitative biomechanical analysis is essential for clinical diagnosis and injury prevention but is often restricted to laboratories due to the high cost of optical motion capture systems. While multi-view video approaches have lowered barriers, they remain impractical for home-based scenarios requiring monocular capture. This paper presents SAM4Dcap, an open-source, end-to-end framework for estimating biomechanical metrics from monocular video without additional training. SAM4Dcap integrates the temporally consistent 4D human mesh recovery of SAM-Body4D with the OpenSim biomechanical solver. The pipeline converts reconstructed meshes into trajectory files compatible with diverse musculoskeletal models. We introduce automated prompting strategies and a Linux-native build for processing. Preliminary evaluations on walking and drop-jump tasks indicate that SAM4Dcap has the potential to achieve knee kinematic predictions comparable to multi-view systems, although some discrepancies in hip flexion and residual jitter remain. By bridging advanced computer vision with established biomechanical simulation, SAM4Dcap provides a flexible, accessible foundation for non-laboratory motion analysis. The code and demo are available at: \href{https://github.com/wanglihx/SAM4Dcap-core}{GitHub}\end{abstract}

% \begin{IEEEkeywords}
% component, formatting, style, styling, insert
% \end{IEEEkeywords}

\section{Introduction}
Quantitative biomechanical analysis is critical for decoding human motion, musculoskeletal loads, and neuromuscular control\cite{ref1,ref2,ref3}. By informing the clinical management of musculoskeletal and neurological conditions, it facilitates accurate diagnosis, injury prevention, tailored rehabilitation, and optimized sports performance\cite{ref4,ref5}.However, biomechanical analysis has long been constrained by a trade-off between accuracy and accessibility. On one hand, marker-based optical motion capture (MoCap) and force plate systems remain the gold standard\cite{ref6,ref7}, yet their prohibitive equipment costs and complex data acquisition protocols often confine them to laboratory settings. On the other hand, emerging video-based technologies such as OpenCap\cite{ref1} have lowered the barrier to entry by demonstrating the feasibility of end-to-end biomechanical analysis in non-laboratory environments. Nevertheless, these methods still necessitate synchronized multi-view or binocular capture, which limits their applicability in many real-world clinical and home-based scenarios where only monocular video is typically available. 

Significant progress has been made in the field of image-based monocular 3D human pose estimation\cite{ref7,ref8,ref9}. Human Mesh Recovery (HMR) methods aim to reconstruct 3D human geometry from monocular images. Recent approaches such as SAM 3D Body\cite{ref10} have demonstrated high accuracy and robustness across diverse scenes and perspectives. Furthermore, the extension of SAMbody-4D toward temporal consistency enables monocular 3D pose estimation to naturally adapt to video sequences on a training-free basis\cite{ref11}. This allows for the consistent modeling of continuous motion. However, these methods primarily focus on kinematic reconstruction and visualization, and their integration with biomechanical analysis remains limited.
% We conclude that with the increasing of prompt points, the zero-shot performance may be on par with the box-prompt mode. 

Building upon prior research, we present SAM4Dcap (Training-free Clinical Biomechanics from Monocular Video), a training-free framework for clinical biomechanical analysis based on monocular video. At its core, SAM4Dcap utilizes a 4D human digital twin driven by SAM 3D Body, integrating monocular 3D human reconstruction with the established OpenSim biomechanical solver\cite{ref12}. This integration enables an end-to-end pipeline from video to comprehensive biomechanical results. The framework requires no additional training data, features a modular design, and supports different musculoskeletal models. 

The primary contributions of this work are summarized as follows:
\begin{itemize}
  \item We propose a fully open-source, training-free end-to-end pipeline for monocular video biomechanical analysis, enabling the estimation of kinetic metrics in the absence of force plates and wearable devices.
  \item Our pipeline supports different OpenSim musculoskeletal models, including the BSM model used by SMPL2AddBiomechanics\cite{ref13} and the LaiUhlrich2022 model used by OpenCap\cite{ref1}, thereby accommodating different biomechanical hypotheses.
  \item We enable direct comparison between monocular and multi-view biomechanical results within a coordinate system and implement a lightweight manual keypoint selection tool for optimization. Additionally, we provide a Linux-native build of AddBiomechanics to support efficient local batch solving of inverse kinematics.
\end{itemize}

\section{Related Works}
\subsection{Monocular Video-based Biomechanical Analysis}
Recent advancements have focused on embedding anatomical constraints into deep learning architectures. For instance, HSMR\cite{ref14} employs a Transformer model to learn the mapping from monocular images to SKEL biomechanical parameters by generating pseudo ground truth labels. BioPose\cite{ref7} extracts mesh vertices as "virtual markers" and utilizes a spatio-temporal network, NeurIK, to solve inverse kinematics. A common limitation among these approaches is the scarcity of high-quality annotated training data and the lack of comprehensive dynamics analysis. While MonoMSK\cite{ref15} achieves both kinematic and dynamic analysis by training on datasets with kinetic annotations, its code and model remain closed-source. In summary, existing methods rely heavily on data augmentation and supervised training paradigms.
\subsection{Human Mesh Recovery (HMR)}
Human Mesh Recovery from monocular imagery has witnessed significant progress, with numerous studies enabling 3D mesh estimation from a single 2D image\cite{ref7,ref8,ref9}. The most widely adopted model is SMPL, which intertwines skeletal structure and soft tissue mass within a shape space, a characteristic that may compromise its anatomical plausibility\cite{ref10,ref14}. To address this, SMPL2AddBiomechanics integrates mesh reconstruction with biomechanical analysis by mapping SMPL-based representations onto musculoskeletal models. Furthermore, SAM 3D Body achieves mesh recovery based on the MHR model, enhancing the anatomical accuracy of the reconstructed geometry\cite{ref10}.
% The emergence of LLM has led to a shift in the research goals of many AI researchers toward the development of large or foundation models. This trend has gained immense popularity and is currently one of the hottest topics in AI research. One viewpoint in the current academic community is that emergence capability can be achieved when the model parameters reach a sufficiently large level, resulting in impressive intelligent processing capabilities. As a result, many AI researchers are considering whether the same is true in the CV domain and looking forward to developing CV foundation models. In this context, segmenting anything becomes an imperative foundation model. Recently, two major parallel works were opened, one is SAM~\cite{kirillov2023segment} and the other is "Segment Everything Everywhere All at Once" named SEEM~\cite{zou2023segment}. During the past two weeks, SAM had a greater degree of impact, but SEEM showed that it has many aspects of performance that outperform SAM and also deserve our attention, measurement, and secondary development.

\section{Method}
\subsection{Problem Definition and Overview}\label{AA}
SAM4Dcap aims to reconstruct a time-consistent 3D human representation and convert it into a Biomechanical Twin compatible with OpenSim-based solvers, finally outputting kinematics and dynamics results. As shown in the Figure.1, there are three main components in SAM4Dcap system:
\begin{itemize}
  \item SAM-Body4D\cite{ref11} with automated prompts: training-free 4D human mesh recovery from monocular video with auto prompts;
  \item HMR-to-TRC Convertor: converts MHR parameters to OpenSim-style marker trajectories (TRC) under a consistent world coordinate system;
  \item Biomechanical Solver: (OpenCap or AddBiomechanics backend) performs kinematics and dynamics results.
\end{itemize}
\subsection{SAM-Body4D with automated prompts}

\begin{figure*}
    \centering
    \includegraphics[width=1\linewidth]{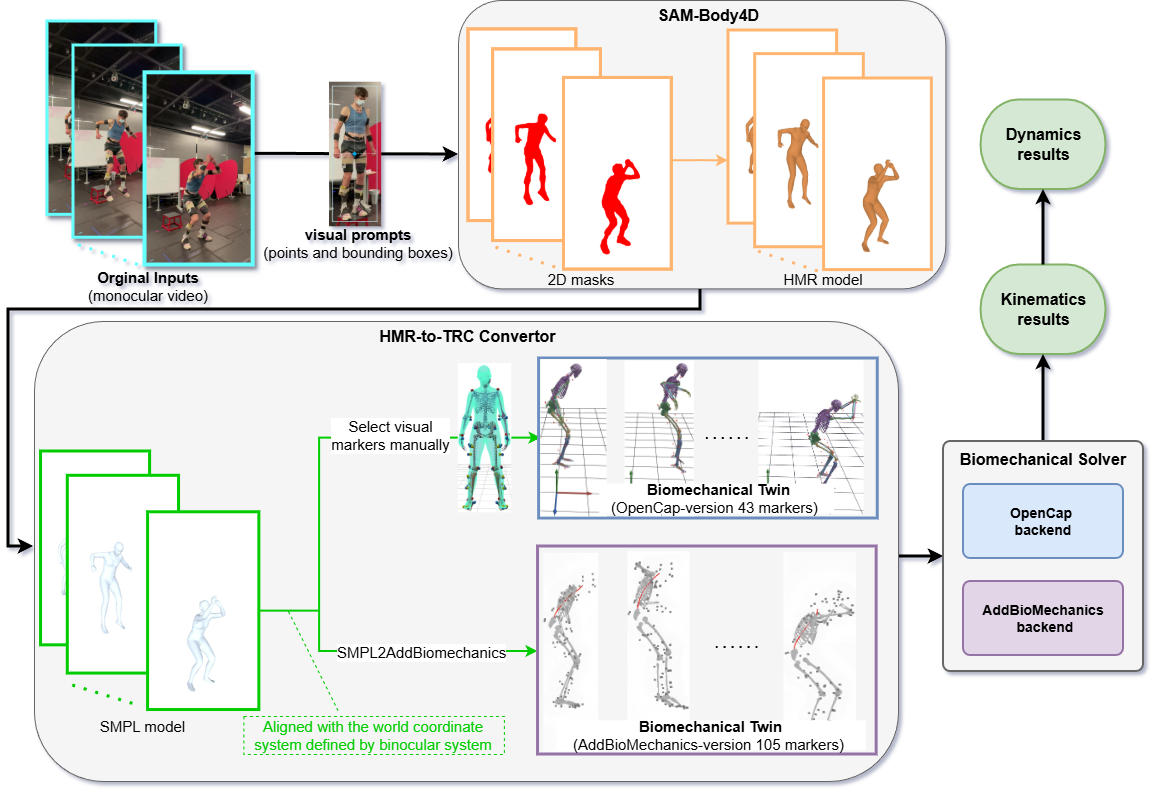}
    \caption{Overall workflow of SAM4Dcap}
    \label{fig:placeholder}
\end{figure*}

SAM-Body4D \cite{ref11} is a training-free framework for temporally consistent 4D human mesh recovery. Building upon SAM-Body4D, we employ automated prompting using a combination of points and bounding boxes.

% Given a monocular video sequence \mathcal{V}\text{={}I_t{{\}}_{t=1}^T}, we employ a ViTDet based human detector to localize the target individual in the initial frame. The detector \mathcal{D} outputs the bounding box with the highest confidence score and confidence threshold {τ_{det}} was set at 0.5.

Given a monocular video sequence $\mathcal{V}=\{I_t\}_{t=1}^{T}$, we employ a ViTDet-based human detector to localize the target individual in the initial frame. The detector $\mathcal{D}$ outputs the bounding box with the highest confidence score, and the confidence threshold $\tau_{\mathrm{det}}$ was set to $0.5$.
\[
\{(\mathbf{b}_i, s_i)\}_{i=1}^{N} = \mathcal{D}(I_1), \quad s_i > \tau_{\mathrm{det}} .
\]
For each identified subject, we construct a visual prompt consisting of a box $\mathbf{b}_i$ and a point $\mathbf{p}_i$.
\[
\mathcal{P}_i = \bigl((\hat{\mathbf{b}}_i, \ell_b), (\mathbf{p}_i, \ell_f)\bigr).
\]
A foreground point label $\ell_f$ located at point $\mathbf{p}_i$, which corresponds to the centroid of the bounding box $\mathbf{b}_i$. The bounding box $\mathbf{b}_i$ itself, accompanied by its box label $\ell_b$.
Figure.2 showed the visual prompt.

\begin{figure}[ht]
    \centering
    \includegraphics[width=\columnwidth]{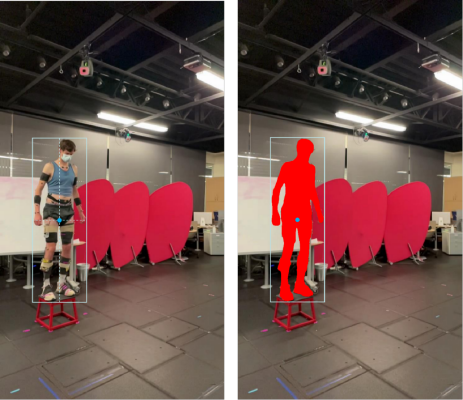}
    \caption{Automated Prompt}
    \label{fig:placeholder}
\end{figure}
\subsection{How to properly prompt the SAM with boxes and points}

Following the SAM-Body4D pipeline, a video segmentation model $S$ (SAM 3) is employed to propagate the initial segmentation mask throughout the entire sequence. Guided by the visual prompt $P_i$ generated in the first frame, the model produces temporally consistent masks through iterative propagation:
\[
M_t = S\!\left(I_t, M_{t-1}, \mathcal{P}\right), \; t = 2, \ldots, T.
\]
The remaining stages of the SAM-Body4D pipeline are subsequently executed to reconstruct a temporally consistent MHR sequence. These meshes serve as the foundational inputs required for the succeeding module.

\subsection{HMR-to-TRC Convertor}\label{AA}

The HMR-to-TRC Converter transforms the reconstructed MHR sequences into OpenSim compatible TRC (Track Row Column) files. We employ the official Meta conversion tool (https://github.com/facebookresearch/MHR) to transform MHR sequences to SMPL sequences. The conversion uses vertex-based optimization fitting with shared body shape parameters across all frames to ensure temporal consistency.As a mainstream open-source library for biomechanical analysis, OpenSim enables motion analysis by simulating the dynamic processes of muscle-driven skeletal joints. SMPL2AddBiomechanics and OpenCap define OpenSim musculoskeletal models with 105 and 43 markers, respectively. Specifically, the former maintains an explicit index mapping relationship with the 6,890 vertices of the SMPL model. In contrast, OpenCap relies on multi-view videos via a pre-trained LSTM enhancer to predict markers from 20 sparse keypoints, thus lacking a direct topological association with SMPL vertices.
To achieve a unified definition of markers across different models, the pre-defined mapping relationship is retained for SMPL2AddBiomechanics. For OpenCap, a mapping strategy is proposed to unify marker definitions across models. First, an initial semantic matching is performed based on anatomical names for the overlapping points between the 105-point SMPL2AddBiomechanics and the 43-point OpenCap sets. Subsequently, the SMPL mesh is aligned with OpenCap marker coordinates to manually identify anatomical regions where name-based matching proves unreliable. By reviewing and correcting these anatomical positional deviations, the reliability of extracting OpenCap markers directly from the SMPL mesh is ensured.
To establish the correspondence between SMPL vertices and OpenCap marker positions, we employ a semi-automatic approach:
Step 1: Anchor-based Procrustes Alignment. Procrustes analysis is a classical shape alignment method that finds the optimal similarity transformation (scale, rotation, translation) to minimize the distance between two sets of corresponding points. Using 7 anatomical anchors, we align the SMPL mesh to OpenCap marker coordinates:
\[
v_{\mathrm{aligned}} = s \cdot R \, v_{\mathrm{smpl}} + t,
\]
where $v_{\mathrm{smpl}} \in \mathbb{R}^3$ denotes the SMPL mesh vertex coordinates,
$v_{\mathrm{aligned}} \in \mathbb{R}^3$ is the transformed aligned vertex,
$s \in \mathbb{R}^{+}$ is the scale factor,
$R \in SO(3)$ (the special orthogonal group in 3D, i.e., rotation matrices satisfying $R^{\top}R = I$ and $\det(R)=1$) is the rotation matrix,
and $t \in \mathbb{R}^3$ is the translation vector.
These parameters are obtained by minimizing:

\[
\min_{s,\mathbf{R},\mathbf{t}} \sum_{i=1}^{7}
\left\lVert s \cdot \mathbf{R}\,\mathbf{p}^{\mathrm{smpl}}_{i}
+ \mathbf{t} - \mathbf{p}^{\mathrm{opencap}}_{i} \right\rVert^{2},
\]
where $\mathbf{p}^{\mathrm{smpl}}_{i}$ and $\mathbf{p}^{\mathrm{opencap}}_{i}$ are the positions of the $i$-th anchor in the SMPL mesh and the OpenCap marker set, respectively.
Step 2: Interactive Refinement. Using a visualization tool, we manually select 16 markers (8 per side) for anatomical regions where automatic matching is unreliable: 5th metatarsal, toe, three shank markers, and three thigh markers.
Step 3: Symmetry Mapping. Left-side markers are derived from right-side selections using SMPL mesh symmetry indices.

For each frame, marker positions are transformed from the camera coordinate system to the world coordinate system defined by the OpenCap binocular pipeline to facilitate subsequent comparisons between monocular and binocular results.
\[
\mathbf{p}_{\mathrm{world}}
=
\
\mathbf{R}_{\mathrm{cam}}^{\mathsf T}
\left(
\mathbf{p}_{\mathrm{cam}} \cdot s_{\mathrm{height}} \
-
\mathbf{t}_{\mathrm{cam}}
\right),
\]
where $\mathbf{p}_{\mathrm{cam}} \in \mathbb{R}^{3}$ is the marker position in camera coordinates (meters),
$\mathbf{p}_{\mathrm{world}} \in \mathbb{R}^{3}$ is the marker position in world coordinates (meters),
$\mathbf{R}_{\mathrm{cam}} \in SO(3)$ is the camera rotation matrix,
$\mathbf{t}_{\mathrm{cam}} \in \mathbb{R}^{3}$ is the camera translation vector (millimeters), and
$s_{\mathrm{height}} = h_{\mathrm{subject}}/h_{\mathrm{predicted}}$ is the height-based scale factor computed from the subject's known height $h_{\mathrm{subject}}$ and the predicted mesh height $h_{\mathrm{predicted}}$.
%%%%%
A vertical offset is applied to ensure the lowest marker point is at ground level:
\[
\mathbf{p}_{\mathrm{final}}=\mathbf{p}_{\mathrm{world}}+\begin{bmatrix}0\\ \Delta y\\ 0\end{bmatrix},
\qquad
\Delta y=-\min_{t}\left(\min_{i}\left(p^{t}_{y,i}\right)\right),
\]
where $p^{t}_{y,i}$ denotes the vertical (y-axis) coordinate of the $i$-th marker at frame $t$.

The final 43-point and 105-point marker trajectories, aligned to the world coordinate system, are exported in TRC format to ensure compatibility with OpenSim.

For futher calculation, this TRC need to be aligned to the world coordinate system expected by binocular system like OpenCap or AddBiomechanics (two optional pipelines for two different solver backend). For the OpenCap pipline, we need a 43-marker set compatible with OpenPose definition. To facilitate this process, we developed an interactive annotation tool, users click 43 marks on a standing (T-pose or neutral standing) frame to match OpenPose keypoints. For the AddBiomechanics pipeline, the library’s marker mapping(for 105 virtual markers) can be applied directly, and alignment uses the model’s built-in conventions. Additionally, we have implemented a Linux-compatible visualization tool for AddBiomechanics (not originally supported by the library), which significantly aids in matching result verification. Once the model is matched from any one of this two piplines, a Biomechanical Twin ready for futher biomechanical calculation is created. 

\subsection{Biomechanical Solver}\label{AA}
The source code repositories of OpenCap (for 43 markers) and AddBiomechanics (for 105 markers) are directly utilized to perform inverse kinematics (IK) analysis. While the original AddBiomechanics source code was compiled for Windows, it has been recompiled in this study to ensure compatibility with the Linux-based research equipment. Additionally, the opencap-process package supports further dynamic analysis beyond the IK results.

\subsection{Methods Summary}\label{AA}
The overall SAM4Dcap pipline can be express as pseudo code and Figure.3 showed the workflow.

\begin{figure}
    \centering
    \includegraphics[width=\columnwidth]{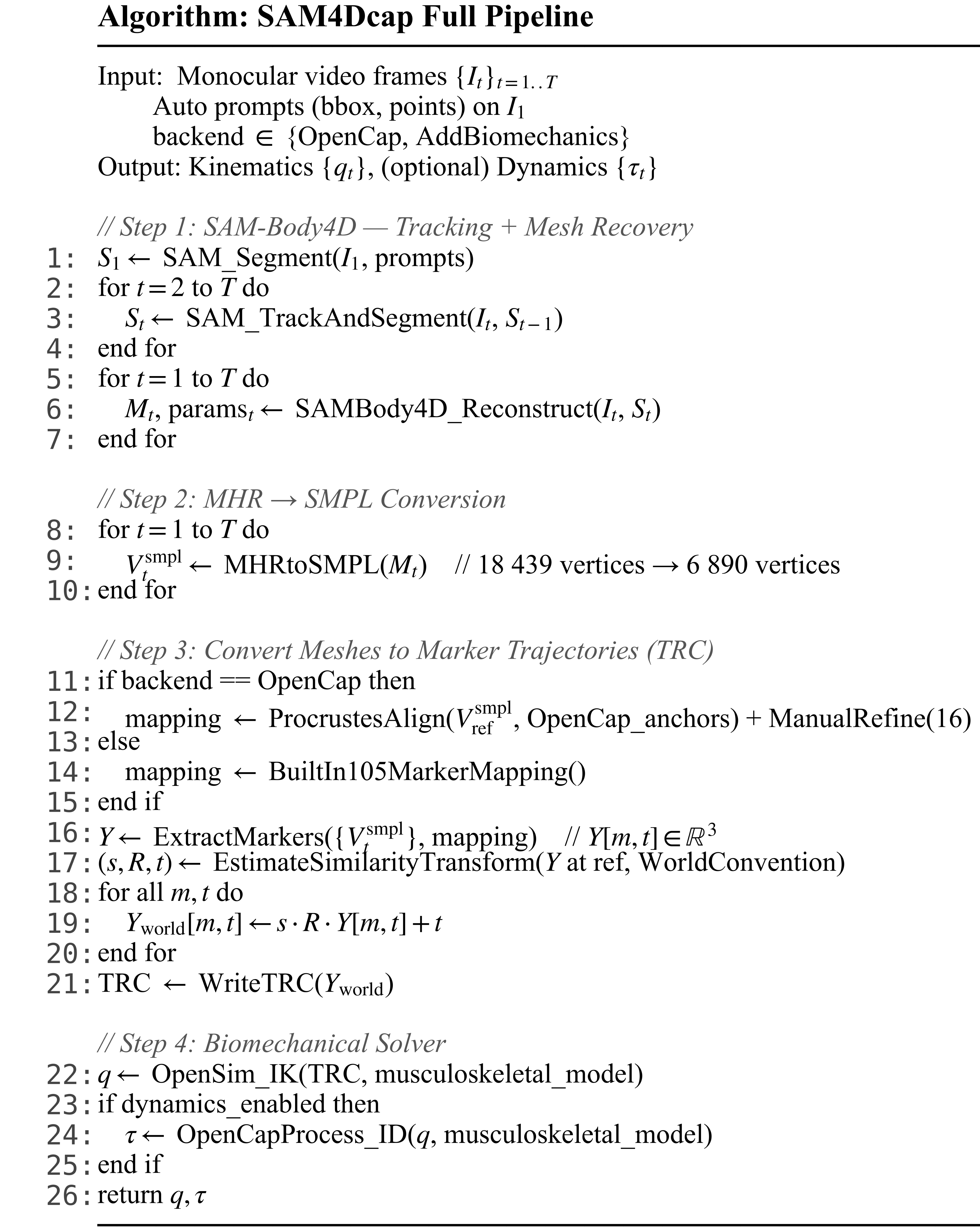}
    \caption{Workflow pseudo code}
    \label{fig:placeholder}
\end{figure}

\section{Results}
This is a preliminary study, and the results are presented as a GitHub repository demonstration and a single-case analysis. Figures 4-7 show the kinematic results of the bilateral hips and knees obtained with MoCap, OpenCap, and SAM4Dcap during one walking session and a drop-jump task.

\begin{figure*}
    \centering
    \includegraphics[width=0.8\linewidth]{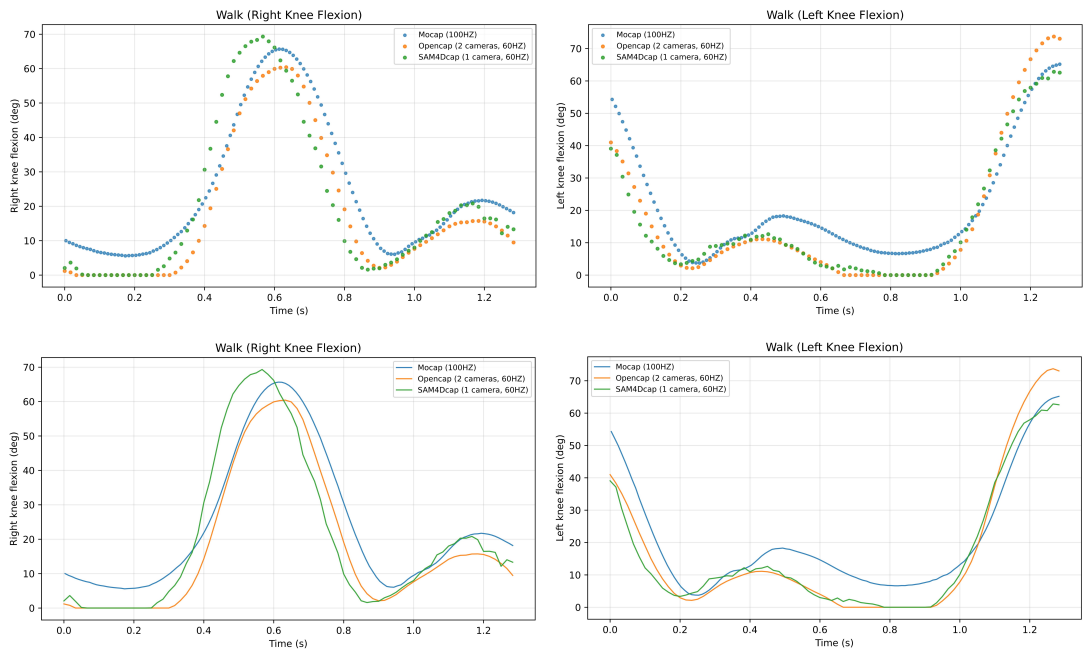}
    \caption{Walk (Knee)}
    \label{fig:placeholder}
\end{figure*}

\begin{figure*}
    \centering
    \includegraphics[width=0.8\linewidth]{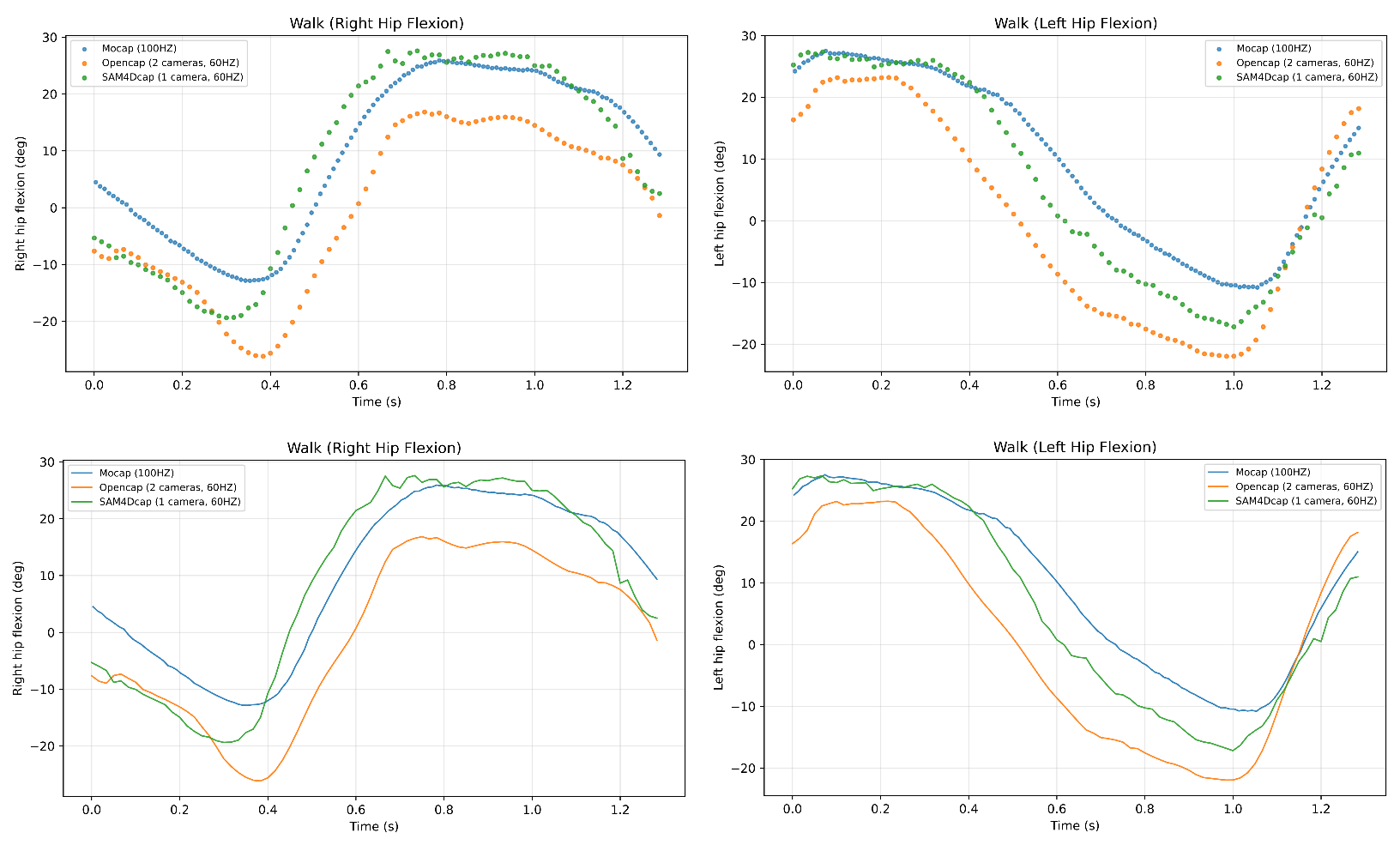}
    \caption{Walk (Hip)}
    \label{fig:placeholder}
\end{figure*}

Under a training-free setting, qualitative analysis indicates that SAM4Dcap achieves knee joint kinematic predictions comparable to, and in some cases visually better than, OpenCap for both movements using the LaiUhlrich 2022 model. For walking, the predicted hip kinematics appear closer to those from MoCap. However, during the drop-jump task, substantial errors are observed at peak hip flexion: left hip $79.63^\circ$ ($-22.23^\circ$) and right hip $79.46^\circ$ ($-20.55^\circ$).

\begin{figure*}
    \centering
    \includegraphics[width=0.8\linewidth]{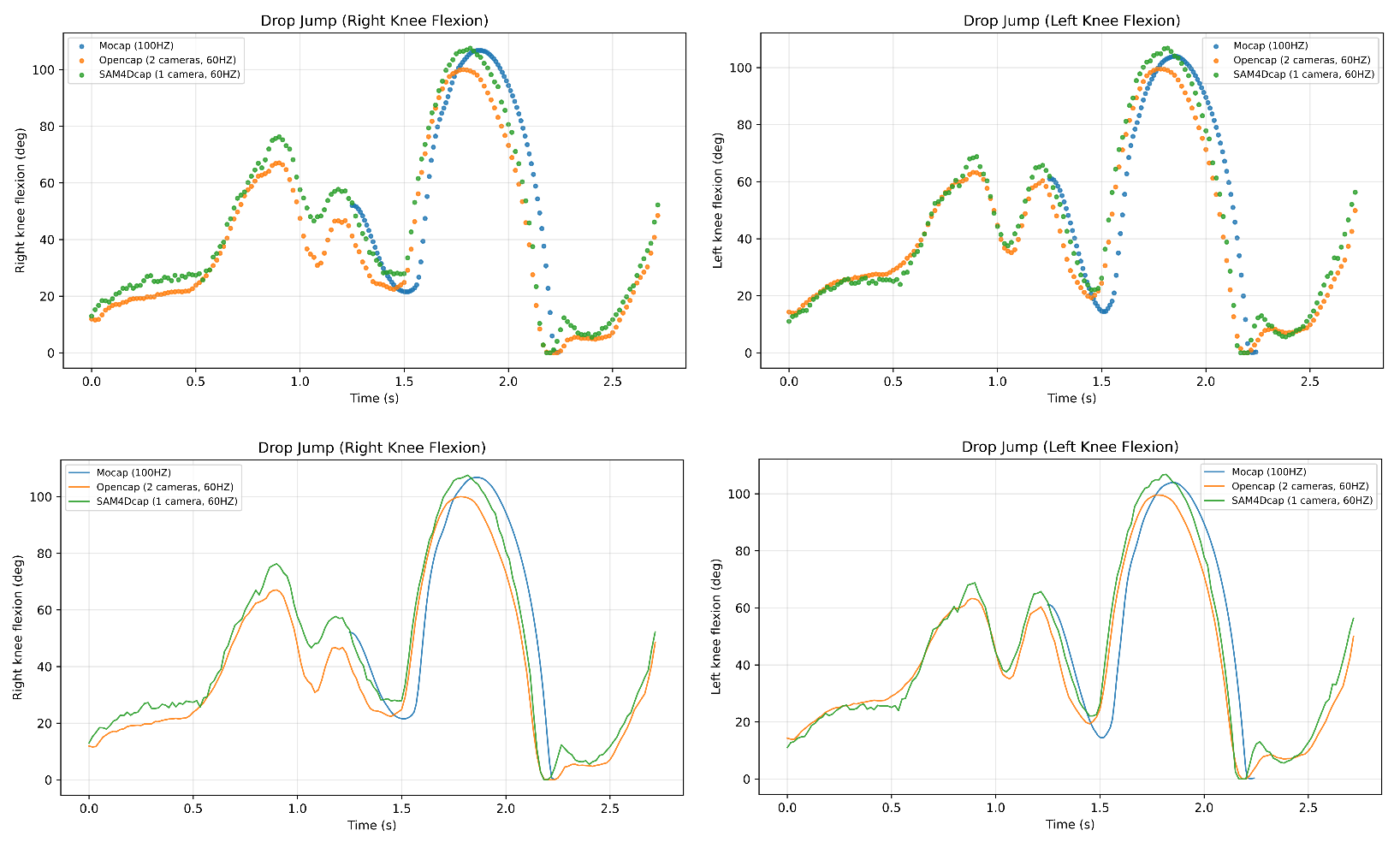}
    \caption{Jump (Knee)}
    \label{fig:placeholder}
\end{figure*}

\begin{figure*}
    \centering
    \includegraphics[width=0.8\linewidth]{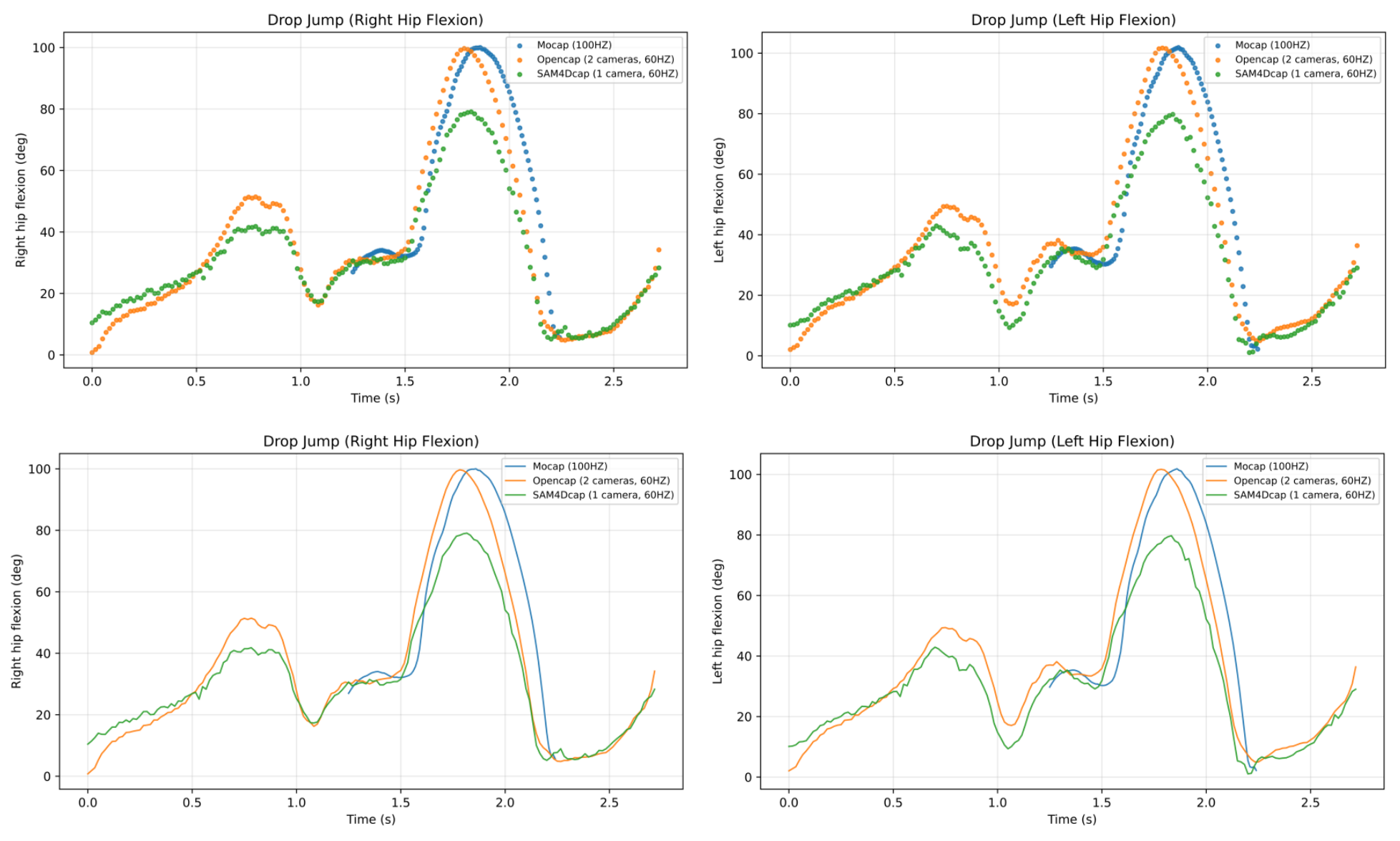}
    \caption{Jump (Hip)}
    \label{fig:placeholder}
\end{figure*}

In addition, the line plots reveal residual jitter in the SAM4Dcap results, which may be attributable to the direct reuse of SAM-Body4D without additional temporal smoothing.

\section{Discussion}
This study does not introduce additional algorithmic designs. Instead, it integrates existing resources to demonstrate the feasibility of performing biomechanical analysis from monocular video. The entire pipeline is flexibly adaptable to different backends. In addition to OpenCap and AddBiomechanics used in this project, the recently released biomechanical diffusion model GaitDynamics\cite{ref5} can also be integrated into this framework. The kinematic results produced in this study can be further used as inputs for GaitDynamics.

This training-free pipeline can be further optimized in future work. First, additional refinement and smoothing can be applied within the SAM-Body4D process to reduce jitter. Second, in order to reuse existing resources, this study converted the MHR model to an SMPL model. This step may reduce biomechanical accuracy because SMPL simplifies human joints into generic ball-and-socket joints. Such a skeleton is physically inaccurate and inherently leads to less accurate pose estimation. In contrast, the MHR model is more anatomically plausible. 

In this study, a manual keypoint selection strategy was adopted for coarse landmark matching. For HMR-based models, while the indices of skin vertices and anatomical joints remain consistent across different subjects, estimating the precise offset between soft-tissue markers and the underlying bone remains a challenge. Furthermore, this distance varies significantly with individual body morphology. In our future work, personalized subject-specific scaling could be implemented based on the MHR model. This would follow a methodology similar to SMPL2AddBiomechanics, utilizing the OSSO model\cite{ref13} to calculate the positions of skin markers relative to the bone surfaces. All subsequent large-scale and batch evaluations will be conducted based on these optimizations.

\section{Conclusion}
Overall, by building upon SAM 3D Body and SAM-Body4D, we further extend their application boundaries and establish a foundational framework for efficient and flexible biomechanical analysis from monocular video in future research.

\end{document}